\pdfoutput=1
\documentclass[11pt]{article}

\usepackage[final]{coling}

\usepackage{times}
\usepackage{latexsym}

\usepackage[T1]{fontenc}
\usepackage[utf8]{inputenc}

\usepackage{microtype}

\usepackage{inconsolata}

\usepackage{graphicx}
\usepackage{multirow}

\usepackage{makecell}
\usepackage{float}

\title{BBPOS: BERT-based Part-of-Speech Tagging for Uzbek}

\author{
 \textbf{Latofat Bobojonova\textsuperscript{1}}\quad
 \textbf{Arofat Akhundjanova\textsuperscript{2}}\quad
 \textbf{Phil Ostheimer\textsuperscript{1}}\quad
 \textbf{Sophie Fellenz\textsuperscript{1}}\quad
\\
 \textsuperscript{1}RPTU Kaiserslautern-Landau\quad
 \textsuperscript{2}Saarland University
\\
 bobojono@rptu.de\quad 
 arak00001@stud.uni-saarland.de\\
 \{ostheimer, fellenz\}@cs.uni-kl.de
}

\begin{document}
\maketitle
\begin{abstract}
This paper advances NLP research for the low-resource Uzbek language by evaluating two previously untested monolingual Uzbek BERT models on the part-of-speech (POS) tagging task and introducing the first publicly available UPOS-tagged benchmark dataset for Uzbek. Our fine-tuned models achieve 91\% average accuracy, outperforming the baseline multi-lingual BERT as well as the rule-based tagger. Notably, these models capture intermediate POS changes through affixes and demonstrate context sensitivity, unlike existing rule-based taggers. 
\end{abstract}

\section{Introduction}
\label{sec: intro}

Uzbek (a.k.a Northern Uzbek) is the second most-spoken language among all Turkic languages after Turkish \citep{johanson_turkic_2015}. It has approximately 40 million native speakers and is the official language of the Republic of Uzbekistan. Although the official script for Uzbek is Latin, for historical reasons, it still heavily relies on Cyrillic script, both unofficially and officially. Uzbek is a morphologically rich language (MLR) and ranks as one of the most agglutinative languages in the world. 

Although Uzbek is a low-resource language, several language models, particularly BERT-based models, have been pre-trained for Uzbek in recent years (e.g. \citealp{mansurov_uzbert_2021}; \citealp{Mamasaidov2023TahrirchiBERT}; \citealp{davronov2024uzroberta}; \citealp{kuriyozov-etal-2024-bertbek}). These models vary in size, quality, and the script of the data on which they have been pre-trained. While some are community projects rather than formal academic publications and lack comprehensive evaluation, others have been assessed only in terms of Masked Language Modeling (MLM) accuracy, with comparisons to multilingual mBERT \citep{DevlinCLT19}. This limitation stems from the lack of publicly available benchmark datasets for Uzbek \citep{mansurov_uzbert_2021}. The main goal of this paper is to fill this gap by creating a new dataset for a downstream task and evaluating models based on this benchmark.

One such downstream task is POS tagging, which lacks publicly available annotated datasets or pre-trained models for Uzbek. POS tagging, specifically with neural models, has the potential to impact linguistic analysis, corpus linguistics, and computational efficiency \citep{allaberganova_neural_2023}. Existing rule-based solutions lack context sensitivity, a limitation that a BERT model can address effectively through its attention mechanism \citep{MuratA24}.
Finally, the fine-tuning approach using pre-trained language models may be the most effective solution for low-resource languages, helping to bridge both the resource and accuracy gap.

In this work, we introduce the first BERT-based POS tagging models (BBPOS) for Uzbek, available for two actively used scripts, Latin and Cyrillic, together with a newly POS-tagged dataset of 500 sentences. Our models show an average accuracy of 91\% based on 5-fold cross-validation.

\section{Related Work}
\label{sec: related-work}

\paragraph{Rule-Based POS Taggers:} \citet{sharipov_uzbektagger_2023} present UzbekTagger — a rule-based POS tagger tool that tags a word by looking up its root form from the dictionary. When it fails to find it, the tagger refers to the neighbouring words to make a decision using six custom grammatical rules. However, the tool only considers the immediate context, making it inferior to neural models (see Section \ref{sec: results}).

\paragraph{Statistical POS Taggers:} \citet{elov_pos_2023} demonstrate the application of Hidden Markov Models (HMMs) on Uzbek by manually tagging a small set of sentences, without developing a full model or dataset.

\paragraph{Neural POS Taggers:} \citet{MuratA24} present the only work on neural POS tagging in Uzbek, alongside two other MRLs: Uyghur and Kyrgyz. The authors propose a new POS tagging method for MRLs using a deeper representation through affix embeddings. They also employ a multi-head attention mechanism to the baseline models and capture dependencies between words regardless of their distance, thereby addressing POS tag ambiguity. This approach achieved an overall accuracy of 79.74\% for Uzbek, representing an increase of up to 4.13\% over other models that utilize only BiLSTMs, CNNs, and CRFs. Unfortunately, their trained models are not publicly available.

\paragraph{Dataset \& Tagset:}
Initial work on the Uzbek morphological tagset identified 12 POS tags that correspond to word classes in traditional Uzbek grammar \citep{abjalova_methods_2021}. \citet{sharipov_uzbektagger_2023} applied this tagset, though their annotated dataset has not been made publicly available. \citet{MuratA24} used a distinct set of 12 POS labels in their dataset designed to be suitable for Uzbek, Uyghur and Kyrgyz. Although the dataset is relatively large, with 20k sentences in the training set and over 23k distinct stems in the Uzbek corpus, it is not publicly available. More morphologically comprehensive tagsets with over 100 tags were also proposed by \citet{sharipov_creating_2022} and \citet{abdullayeva_o_2022}, but no tagged datasets based on these frameworks currently exist.
\section{Experiments}
\label{sec: experiments}

\subsection{Methods} 
Due to the lack of a public dataset for POS tagging, we created our own dataset\footnote{The dataset is publicly available at \url{https://huggingface.co/datasets/latofat/uzbekpos}} (see Section \ref{sec: tagset-dataset}). We chose one pre-trained model for each script (see Section \ref{sec: models}) and fine-tuned them\footnote{One fine-tuned model per script is available at: \url{https://huggingface.co/latofat}} with our dataset for the POS tagging task. As a baseline, we fine-tuned a multi-lingual mBERT model \citep{DevlinCLT19}. Each type of model was individually evaluated using a 5-fold cross-validation with a 80\% - 20\% train-test split. All BERT models were fine-tuned with the same hyperparameters (see Appendix \ref{appendix: hyperparams}).  
\begin{table}
\centering
\begin{tabular}{|c|l|r|r|}
\hline
\textbf{Index} & \textbf{POS tag} & 
\textbf{\makecell{\# of \\ words}} & 
\textbf{\makecell{\# of unique \\ words}}\\\hline
0 & ADJ     &   454     &   356     \\
1 & ADP     &   189     &   48      \\
2 & ADV     &   152     &   102     \\
3 & AUX     &   96      &   27      \\
4 & CCONJ   &   85      &   7       \\
5 & DET     &   \textbf{16}      &   \textbf{14}      \\
6 & INTJ    &   \textbf{11}      &   \textbf{6}       \\
7 & NOUN    &   2141    &   1751    \\
8 & NUM     &   217     &   94      \\
9 & PART    &   67      &   14      \\
10 & PRON   &   273     &   112     \\
11 & PROPN  &   300     &   261     \\
12 & PUNCT  &   810     &   19      \\
13 & SCONJ  &   \textbf{9}       &   \textbf{3}       \\
14 & SYM    &   \textbf{1}       &   \textbf{1}       \\
15 & VERB   &   1001    &   721     \\
16 & X      &   \textbf{9}       &   \textbf{3}       \\\hline
& \textbf{Total}  &   5831    &   3488    \\\hline
\end{tabular}
\caption{\label{tab: dataset-stats}
Overview of the distribution of tags in the dataset. Bold numbers highlight relatively underrepresented tags.
}
\end{table}

\subsection{Data}
\label{sec: tagset-dataset}
\paragraph{Tagset Selection:}

We used the Universal Part-of-Speech (UPOS) \citep{nivre_upos_2016}, as it is a multilingual tagset that aims to cover similar linguistic features consistently across languages. Currently, it has been the foundation for 283 treebanks in 116 languages\footnote{\url{https://universaldependencies.org/}} and our dataset is the first work to employ UPOS for Uzbek. There are 17 tags in the UPOS as shown in Table~\ref{tab: dataset-stats}, and Uzbek can use all of them. Furthermore, it is easy to map UPOS to 12 word classes identified in traditional Uzbek grammar (see Appendix \ref{appendix: tag-conversion}). 

\begin{table*}
\centering
\begin{tabular}{|l|c|c|c|c|c|}
\hline
 &\textbf{UzbekTagger} & \multicolumn{2}{c|}{\textbf{mBERT}} & \textbf{TahrirchiBERT} & \textbf{UzBERT}  \\ \cline{3-4}
&rule-based&latin&cyrillic&(latin)&(cyrillic) \\ \hline
\textbf{Accuracy}       &  $75.6 \pm 1.6$   & \boldmath{$86.0 \pm 1.0$} & $80.2 \pm 1.0$ & $90.9 \pm 0.9$    & \boldmath{$91.6 \pm 0.4$}          \\ \hline
\textbf{F1}          &  $57.4 \pm 2.3$   & \boldmath{$77.5 \pm 0.9$} & $68.5 \pm 1.9$ & $85.2 \pm 1.3$    & \boldmath{$86.4 \pm 0.6$}          \\ \hline
\end{tabular}
\caption{Accuracy and F1-score for different POS taggers measured in Mean $\pm$ Standard Deviation (\%).}
\label{tab: model-results}
\end{table*}

\paragraph{Dataset Development:} We collected 500 sentences (5,831 words), 250 sourced from news articles and 250 from fictional books. We manually annotated the data written in Latin script with UPOS tags. Then it was transliterated into a Cyrillic script to fine-tune the Cyrillic model.  Table~\ref{tab: dataset-stats} shows the distribution of tags in the dataset and the number of unique words per POS (more details in Appendix \ref{appendix: annotation}). As the sentences are ordered according to their genre, i.e., fiction and news, the datasets for each script were shuffled with the same seed before a 5-fold split for training and testing.

\subsection{Models}
\label{sec: models}

\paragraph{Latin BERT:} We chose the open source TahrirchiBERT \citep{Mamasaidov2023TahrirchiBERT}, a monolingual RoBERTa \citep{roberta} model pre-trained on Uzbek Latin script. It is trained on large text data extracted from online blogs and scanned books (equivalent to 5B tokens $\approx$ 18.5GB). The dataset is fairly noisy due to the errors introduced by poor OCR applied to the books. Additionally, TahrirchiBERT does not handle the required pretokenization rules for the Latin script of Uzbek. Specifically, the modifier letters\footnote{A modifier letter functions like diacritics, changing the sound-values of the letter it proceeds. Unlike diacritics, they do not combine with the letter.} used in \texttt{o‘} and \texttt{g‘} letters and the glottal stop sign \texttt{’} are treated as delimiter signs that cause incorrect word splits. The authors introduced a normalization specific to Uzbek Latin script, preventing some common spelling errors.

\paragraph{Cyrillic BERT:} We fine-tuned UzBERT \citep{mansurov_uzbert_2021}, a monolingual BERT model \citep{DevlinCLT19} pre-trained solely on Cyrillic scripted Uzbek text. According to the authors, the model is trained on high-quality Cyrillic text data with 142M words ($\approx$ 1.9GB) and has not been evaluated on any downstream tasks due to the lack of public datasets. There are no Uzbek Cyrillic script-specific rules to be applied during the normalization and pretokenization stages, as each letter in the Uzbek Cyrillic alphabet is represented by a single alphabetic character. 

\section{Results}
\label{sec: results}

Table~\ref{tab: model-results} shows accuracy and F1-score for all trained models together with the results obtained from the rule-based UzbekTagger on the POS-converted dataset (see Appendix \ref{appendix: uzbektagger}). It presents the mean and standard deviation for accuracy and F1-score of 5-fold cross-validation. The rule-based POS tagger with an average accuracy of 75\% falls behind all BERT models. Both monolingual models outperform mBERT by a good margin overall. Table~\ref{tab: model-results}, column three shows Latin Uzbek is better represented in mBERT than Cyrillic Uzbek.

\paragraph{UzBERT vs TahrirchiBERT:} Monolingual BERT models, regardless of script, show similarly high accuracies of at least 90\% and F1-scores of at least 84\%. Having been trained on ten times less data, UzBERT has outperformed TahrirchiBERT by a slight margin in both metrics. We hypothesize that this might be due to the data quality used for pre-training and incorrect pretokenization used for Latin scripted text. Especially during inference, when a sentence has to be pretokenized, TahrirchiBERT fails in successfully tagging words written with one of the modifier letters.  

\begin{table}
\centering
\begin{tabular}{|l|c|c|c|c|c|}
\hline
POS& \makecell{tah-\\rir-\\chi\\(lat)}&\makecell{uz-\\bert\\(cyr)}&\makecell{m-\\bert\\(lat)}&\makecell{m-\\bert\\(cyr)}& \makecell{rel.\\freq.}\\\hline
ADJ     & 77.0   &  79.3  & 54.7 &  23.3    & 8.9 \\
ADP     & 92.8   &  88.6  & 88.5 &  63.0    & 3.6 \\
ADV     & 64.3   &  75.4  & 11.1 &  5.7     & 3.6 \\
AUX     & 88.2   &  83.3  & 90.9 &  55.2    & 1.9 \\
CCONJ   & 84.8    &  94.1   & 90.9 &  90.9    & 1.9 \\
\textbf{DET}     & \textbf{0.0}    &  \textbf{0.0}   & \textbf{0.0}  &  \textbf{0.0}     & \textbf{0.3} \\
\textbf{INTJ}    & \textbf{0.0}    &  \textbf{0.0}   & \textbf{0.0}  &  \textbf{0.0}     & \textbf{0.4} \\
NOUN    & 86.1   &  81.6  & 72.2 &  50.9    & 28.2 \\
NUM     & 88.2   &  93.0  & 80.0 &  83.1    & 3.8 \\
PART    & 92.3   &  85.7  & 92.3 &  64.5    & 1.5 \\
PRON    & 76.8   &  84.7  & 77.2 &  70.6    & 5.8 \\
PROPN   & 90.7   &  87.1  & 77.6 &  53.5    & 4.4 \\
PUNCT   & 98.9    &  100.0   & 98.3 &  99.4    & 18.6 \\
\textbf{SCONJ}   & \textbf{0.0}    &  \textbf{0.0}   & \textbf{0.0}  &  \textbf{0.0}     & \textbf{0.0}     \\
\textbf{SYM}     & \textbf{0.0}    &  \textbf{0.0}   & \textbf{0.0}  &  \textbf{0.0}     & \textbf{0.0}     \\
VERB    & 89.8   &  90.7  & 84.3 &  76.4    & 17.2 \\
\textbf{X}       & \textbf{0.0}    &  \textbf{0.0}   & \textbf{0.0}  &  \textbf{0.0}     & \textbf{0.0}     \\\hline
\end{tabular}
\caption{
F1-scores of BBPOS models for each POS tag, including the tags' relative frequency in the evaluation set. Bold entries indicate tags that models failed to learn.
}
\label{tab: tag-results}
\end{table}

\begin{figure*}
    \centering
    \includegraphics[width=\linewidth]{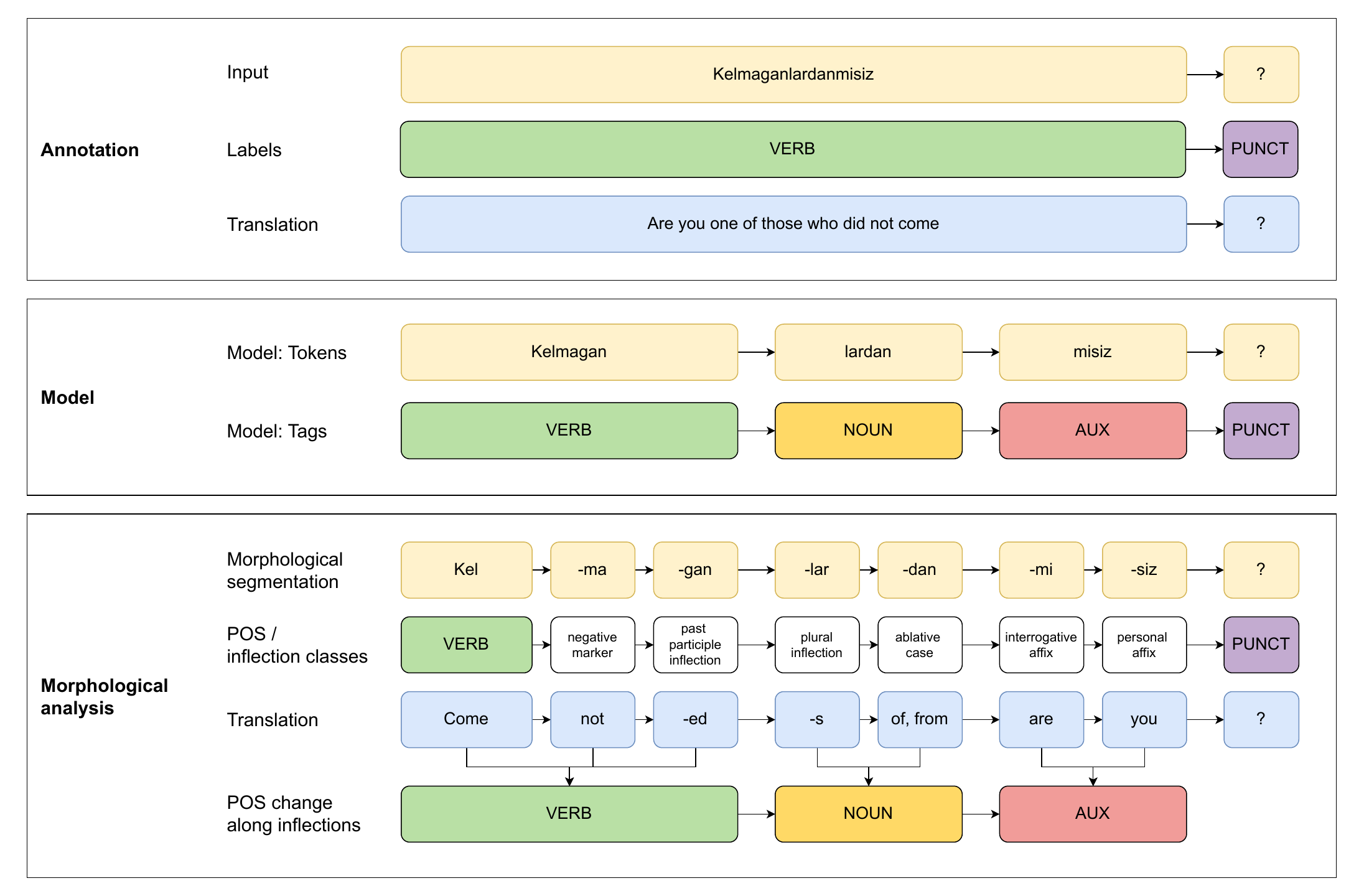}
    \caption{Analysis of one sentence-word in Uzbek: manual annotation according to UPOS guidelines (top); how BBPOS tags it (middle); comprehensive morphological analysis of the word (bottom).}
    \label{fig:kelmaganlardanmisiz}
\end{figure*}

\paragraph{Learning per Tag:} We randomly chose an evaluation fold to evaluate which tags are learned well by the BERT models. In Table~\ref{tab: tag-results}, we present the relative frequency of POS tags in the chosen evaluation fold, together with the F1-scores obtained by the corresponding BERT models that were not trained on it. All models could not learn the same five tags, most likely due to the low representation in the overall dataset (see Table~\ref{tab: dataset-stats}).

\paragraph{Context-sensitivity:} We assess the rule-based and neural models for context sensitivity, running a couple of sentences containing homonyms. 
The sentence \textit{Tortmani tortma} `Don't pull the drawer' should be tagged as \texttt{[NOUN, VERB]}. 
The rule-based UzbekTagger will naturally tag it as \texttt{[NOUN, NOUN]} (treating it as `The drawer drawer'). Similarly, mBERT fails at tagging this same sentence in both Latin and Cyrillic. However, TahrirchiBERT and UzBERT tag it correctly as \texttt{[NOUN, VERB]}.

\section{Discussion}
\label{sec: discussion}
An interesting aspect of our experiments was how our models handled highly inflected words. They learned morphological features by detecting intermediate POS changes through affixes. For instance, in Figure~\ref{fig:kelmaganlardanmisiz} you can see how the word \textit{Kelmaganlardanmisiz?} which corresponds to a whole sentence in English (`Are you one of those who did not come?') is tagged by our models. It also shows manual POS and morphological annotation for it. As you can see, our model's result resembles the morphological analysis rather than the simple POS labeling with which it was trained. In fact, according to Universal Dependencies (UD) guidelines, the word's POS relies solely on its lemma's POS.

Our work on POS tagging has the potential for extension to data generation in morphological analysis, specifically in morpheme classification. However, this requires BERT models to be pre-trained using morphological or morphologically informed tokenizers rather than relying on subword tokenization methods like BPE and WordPiece which are statistical algorithms. Additionally, the success of neural models in learning aspects of Uzbek morphology could inspire the linguistic community to develop a unified and comprehensive POS tagset for Uzbek, one that considers how morphemes influence word-level POS shifts. Previous work on Turkish \citep{coltekin} also discusses the guidelines for this.

The inconsistent representation of the letters \texttt{o‘}, \texttt{g‘} and \texttt{’} in texts, caused by the use of varying forms of apostrophes, poses a significant challenge for Latin Uzbek. This issue, as evidenced by the pre-tokenization problem detected in TahrirchiBERT, underscores the importance of pre-training language models for Latin-scripted Uzbek on data that adheres to consistent alphabet standards. Alternatively, we can focus on pre-training monolingual Uzbek models that apply normalization rules to standardize the singular form of the above letters across diverse Uzbek text data.

%Our quick experiments show that substituting the glottal stop sign \texttt{(’)} with `modifier letter apostrophe' (U+02BC) and modifier letters in \texttt{o‘} and \texttt{g‘} with `modifier letter turned comma' (U+02BB) through normalization prevented incorrect splits in the existing pretokenization algorithms. Future monolingual Uzbek language models in Latin could benefit from our findings and help to sustain this standard.   

\section{Conclusion}
\label{sec: conclusion}
In this work, we introduced a new dataset for the low-resource Uzbek language tagged with the UPOS tagset and trained the first BERT-based POS taggers on it. We evaluated two monolingual Uzbek BERT models on the POS tagging downstream task, identifying potential improvements to pre-train Uzbek language models in the future. Our BBPOS models reached an average accuracy of 91\% on 5-fold cross-validation, outperforming the baseline mBERT and the existing rule-based solution by far both in accuracy and F1-score. They show context sensitivity in handling ambiguous sentences with homonyms. They learned parts of speech for POS changing morphemes, generating enriched annotations with more linguistic information. 

\section*{Limitations}
We acknowledge the following limitations of the fine-tuned models:
\begin{itemize}
    \item Even though our fine-tuned models performed better than the rule-based tagger on the evaluation sets, we acknowledge that our models fail to tag overly inflected words as single tokens due to the subword tokenization used in them. The models can be used for synthetic data generation although with heavy human supervision to ensure quality and accuracy. 

    \item Additionally, due to the poor pretokenization of TahrirchiBERT, the Latin models fail at words containing the letters \texttt{o‘}, \texttt{g‘}, \texttt{’}, as they incorrectly split them into words treating the modifier letters as delimiters. This error is not evident during the validation and training stages of the token classification task as it is during inference. 
\end{itemize}

We also acknowledge the following limitations of our benchmark dataset:
\begin{itemize}
    \item  Our models failed to learn five out of seventeen POS tags due to the small representation in the initial dataset. Our benchmark needs to be enriched on those POS tags.
    \item While not of major importance, our dataset is relatively small. The dataset is insufficient for training POS tagging models from scratch, such as HMM, CRF, RNN, or LSTM. While we trained an HMM model, its poor performance, achieving an accuracy of $(40.7 \pm 1)$ and an F1-score of $(8.9 \pm 1.8)$, proved it to be an inadequate baseline and therefore it is not included in the results.
\end{itemize}

\bibliography{coling}

\appendix

\section{Hyperparameter Settings}
\label{appendix: hyperparams}
Table~\ref{tab: bert-params} shows hyperparameters and their values used in the fine-tuning of BERT models using \texttt{transformers}\footnote{\url{https://huggingface.co/docs/transformers}} library.

For all conducted evaluations we used the sequence labeling evaluation metric -- \texttt{seqeval} -- from the \texttt{evaluate}\footnote{\url{https://huggingface.co/docs/evaluate}} package.

\begin{table}[ht]
\centering
\begin{tabular}{|l|l|}
\hline
learning\_rate & 2e-5 \\\hline
per\_device\_train\_batch\_size & 16 \\\hline
per\_device\_eval\_batch\_size & 16 \\\hline
num\_train\_epochs & 5 \\\hline
weight\_decay & 0.01 \\\hline
\end{tabular}
\caption{Hyperparameters used for fine-tuning the BERT models}
\label{tab: bert-params}
\end{table}

\section{Tagset Conversion}
\label{appendix: tag-conversion}

\begin{table}[ht]
\centering
\begin{tabular}{|c|c|c|}
\hline
\textbf{UPOS} & \textbf{Uzbek POS} & \textbf{Comment}\\\hline
\makecell{NOUN\\PROPN}      & NOUN                      &   \\\hline
\makecell{PRON\\DET}        & PRON                      &   \\\hline
\makecell{CCONJ\\SCONJ}     & CONJ                      &   \\\hline
AUX                         & VERB                      &   \\\hline
ADJ                         & ADJ                       &   \\\hline
ADP                         & AUX                       &   \\\hline
INTJ                        & INTJ                      &   \\\hline
NUM                         & NUM                       &   \\\hline
PART                        & PART                      &   \\\hline
ADV                         & \makecell{MOD\\ADV}       &   \makecell{There are finite\\modal words}\\\hline
VERB                        & \makecell{IMIT\\VERB}     &   \makecell{There are finite\\immitation words}\\\hline
\makecell{PUNCT\\SYM\\X}     & irrelevant                &   \makecell{There is no specific\\POS tag for these\\gorup of tokens in\\the Uzbek grammar}\\\hline
\end{tabular}
\caption{UPOS $\rightarrow$ traditional Uzbek POS}
\label{tab: tag-conversion}
\end{table}

To align UPOS with the traditional Uzbek POS tagset and bridge prior research, we developed a conversion script that maps UPOS tags to Uzbek POS categories. Table~\ref{tab: tag-conversion} shows how individual tags are handled, grouped, or reclassified according to Uzbek linguistic rules \citep{abjalova_methods_2021}. While tags like \texttt{ADJ}, \texttt{INTJ}, \texttt{NUM} and \texttt{PART} are aligned directly, some are merged into broader word classes (e.g. \texttt{PROPN} $\cup$ \texttt{NOUN} = \texttt{NOUN}). 

\begin{figure*}
    \centering
    \includegraphics[width=\linewidth]{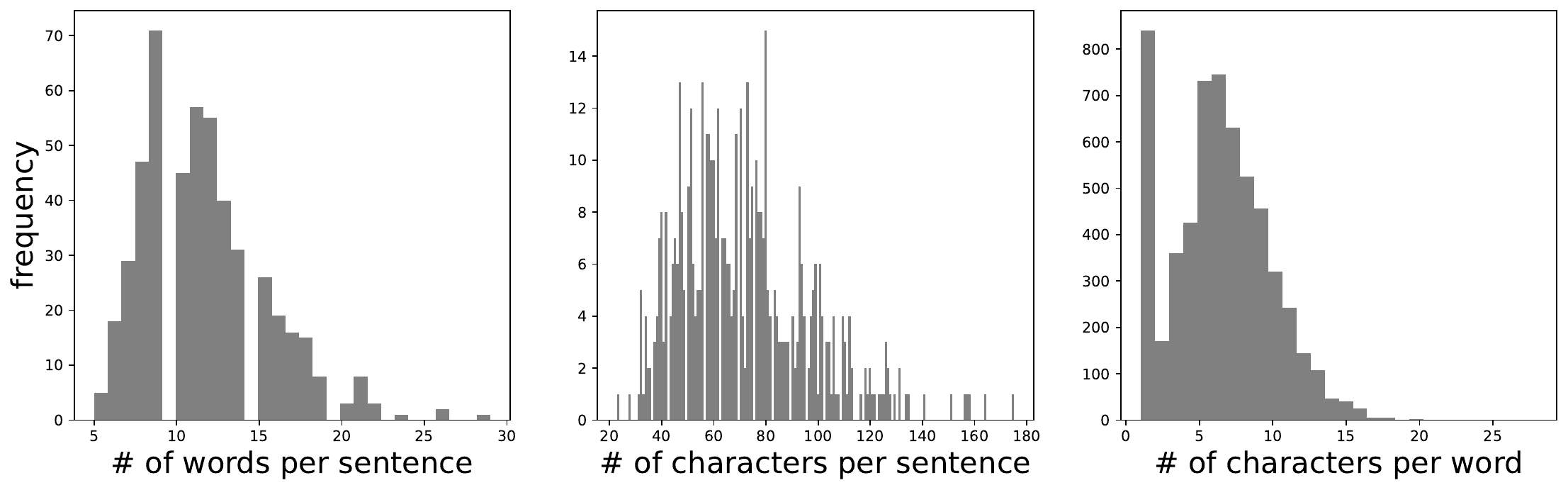}
    \caption{Words per sentence and characters per word/sentence in the dataset.}
    \label{fig: char-word-distributions}
\end{figure*}

The most complex part of this conversion is \texttt{ADV} and \texttt{VERB} tags. Uzbek grammar splits adverbs into action-related (\texttt{ADV}) and attitude-related (\texttt{MOD}). Similarly, \texttt{VERB} is split into true verbs (\texttt{VERB}) and imitative words (\texttt{IMIT}). We made this distinction by simple set membership check, as words that belong to \texttt{MOD} and \texttt{IMIT} classes are finite and uninflected. Moreover, the Uzbek grammar does not specify POS tags for punctuation (\texttt{PUNCT}), symbols (\texttt{SYM}), and miscellaneous categories (\texttt{X}), so we excluded them from the mapping. 

\section{Data Statement}
\label{appendix: annotation}

We chose news and fiction genres to ensure broad domain coverage while preserving diversity in length, formality, and literary quality. All sentences were handpicked to ensure the quality of the data. News texts of the dataset were collected from the major news sites\footnote{\url{https://kun.uz/} and \url{https://daryo.uz/}}. They cover various topics and reflect contemporary Uzbek language use. Fiction texts were chosen from the publicly available Uzbek works on the internet, including: “Og‘riq Tishlar” and “Dahshat” by Abdulla Qahhor, “Shum Bola” and “Yodgor” by G‘afur G‘ulom, “Sofiya”, “Hazrati Hizr Izidan”, “Bibi Salima va Boqiy Darbadar”, “Olisdagi Urushning Aks-Sadosi” and “Genetik” by Isajon Sulton, “Buxoro, Buxoro, Buxoro…”, “Ozodlik” and “Lobarim Mening…” by Javlon Jovliyev, “Ko‘k Tog‘”, “Insonga Qulluq Qiladurmen”, “Fano va Baqo” and “Chodirxayol” by Asqar Muxtor, “Ajinasi Bor Yo‘llar” by Anvar Obidjon, “Kecha va Kunduz” and “Qor Qo‘ynida Lola” by Cho‘lpon.

Figure~\ref{fig: char-word-distributions} shows the number of words per sentence and the number of characters per word/sentence. The number of words per sentence ranges from 5 to 29, with an average of 11–12, likely reflecting natural linguistic patterns in Uzbek. This trend is further illustrated by the average number of characters per sentence (72) and per word (6).

Annotation was performed manually by one of the native Uzbek-speaking authors who is MSc in Computational Linguistics with a background in Uzbek linguistics, applying each UPOS tag according to the Universal Dependencies (UD) guidelines\footnote{\url{https://universaldependencies.org/u/pos/}}. In addition to UD POS tagging guidelines, UD treebanks of other Turkic languages and Uzbek grammar rules \citep{Rahmatullayev} were also used as a point of reference. Ambiguous cases such as the annotation of multiword expressions (MWEs) in compound verbs were solved through extensive discussions with other linguists and UD experts. 

The Latin-scripted dataset was subsequently turned into a morpho-syntactically annotated UD treebank, released as part of UD version 2.15. 

The transliteration was performed using an online transliterator tool\footnote{\url{https://tahrirchi.uz/uz/editor}}.

\section{Comparison with UzbekTagger}
\label{appendix: uzbektagger}
To compare BBPOS models with the rule-based POS tagger tool, we relabeled our golden dataset with 12 conventional Uzbek POS tagset using the conversion script we developed (see Section \ref{appendix: tag-conversion}). The token families that are excluded by the logic of UzbekTagger, such as punctuations, symbols and other (i.e. \texttt{PUNCT, SYM, X}) were eliminated from the dataset to the favor of UzbekTagger results. We then ran the untagged 5 evaluation folds, each containing 100 sentences, through UzbekTagger and compared the results against the relabeled golden dataset.

\end{document}